\documentclass[runningheads]{llncs}
\usepackage[T1]{fontenc}
\usepackage{graphicx}
\usepackage{amsmath}
\usepackage{subcaption}
\usepackage{cite}
\usepackage{booktabs}
\usepackage{hyperref}
\usepackage{color}

\newcommand{\ra}[1]{\renewcommand{\arraystretch}{#1}}
\begin{document}
\title{Pretraining is All You Need: A Multi-Atlas Enhanced Transformer Framework for Autism Spectrum Disorder Classification}
\titlerunning{Pretraining is All You Need: Transformers for ASD Classification}
%

\author{Lucas Mahler\inst{1}\and
Qi Wang\inst{1}\and
Julius Steiglechner\inst{2,1}\and
Florian Birk \inst{1, 2} \and
Samuel Heczko \inst{1} \and
Klaus Scheffler \inst{2, 1}\and
Gabriele Lohmann \inst{2, 1}}
\authorrunning{L. Mahler et al.}
%

\institute{Max-Planck-Institute for Biological Cybernetics, 72076 Tübingen, Germany\and
University Hospital Tübingen, 72076 Tübingen, Germany\\
\email{lucas.mahler@tuebingen.mpg.de}}

\maketitle              
\begin{abstract}
Autism spectrum disorder (ASD) is a prevalent psychiatric condition characterized by atypical cognitive, emotional, and social patterns.
Timely and accurate diagnosis is crucial for effective interventions and improved outcomes in individuals with ASD.
In this study, we propose a novel Multi-Atlas Enhanced Transformer framework, METAFormer, ASD classification.
Our framework utilizes resting-state functional magnetic resonance imaging data from the ABIDE I dataset, comprising 406 ASD and 476 typical control (TC) subjects. 
METAFormer employs a multi-atlas approach, where flattened connectivity matrices from the AAL, CC200, and DOS160 atlases serve as input to the transformer encoder.
Notably, we demonstrate that self-supervised pretraining, involving the reconstruction of masked values from the input, significantly enhances classification performance without the need for additional or separate training data.
Through stratified cross-validation, we evaluate the proposed framework and show that it surpasses state-of-the-art performance on the ABIDE I dataset, with an average accuracy of 83.7\% and an AUC-score of 0.832.
The code for our framework is available at \href{https://github.com/Lugges991/METAFormer}{github.com/Lugges991/METAFormer}
\end{abstract}
\section{Introduction}
Autism spectrum disorder (ASD) is a widespread psychiatric condition characterized by atypical cognitive, emotional, and social patterns.
With millions of individuals affected worldwide, the early diagnosis of ASD is a critical research priority, as it has a significant positive impact on patient outcomes.
The etiology of ASD remains elusive, with intricate interactions among genetic, biological, psychological, and environmental factors playing a role.
Currently, diagnosing ASD relies heavily on behavioral observations and anamnestic information, posing challenges and consuming a considerable amount of time.
Skilled clinicians with extensive experience are required for accurate diagnosis.
However, common assessments of ASD have been criticized for their lack of objectivity and transparency \cite{Timimi2019}.
Given these limitations, there is an urgent need for a fast, cost-effective, and objective diagnostic method that can accurately identify ASD leading to more timely interventions and improved outcomes for affected individuals.

In recent years, magnetic resonance imaging (MRI) has emerged as a powerful non-invasive tool for gaining insights into brain disorders' pathophysiology.
Functional MRI (fMRI), a notable advancement in MRI technology, allows for the investigation of brain function by measuring changes in blood oxygen levels over time.
Functional connectivity (FC) analysis\cite{Biswal1995} plays a crucial role in fMRI data analysis, as it examines the statistical dependencies and temporal correlations among different brain regions.
Rather than considering isolated abnormalities in specific regions, brain disorders often arise from disrupted communication and abnormal interactions between regions.
FC analysis enables researchers to explore network-level abnormalities associated with various disorders.
This analysis involves partitioning the brain into regions of interest (ROIs) and quantifying the correlations between their time series using various mathematical measures.

In recent years machine learning approaches have been widely applied to the problem of ASD classification using resting-state fMRI (rs-fMRI) data.
The majority of these studies use functional connectivities obtained from a predefined atlas as input to their classifiers.
A considerable amount of work used classical machine learning algorithms such as support vector machines and logistic regression to classify ASD\cite{Du2018}.
However, these methods have limitations as they are typically applied to small datasets with specific protocols and fixed scanner parameters, which may not adequately capture the heterogeneity present in clinical data.
3D Convolutional neural networks\cite{warum, Zhao2018, Thomas2020} have also been applied to preprocessed fMRI data, \cite{DarkASD} have used 2D CNNs on preprocessed fMRI data.
Though, these approaches are as well limited by the fact that they were only using small homogeneous datasets.\newline
More recent works tried to overcome the homogeneity limitations and have used deep learning approaches to classify ASD based on connectomes.
Multi-layer perceptrons are suited to the vector based representations of connectomes and have thus seen some usage in ASD classification\cite{aimafe, misodnn}.
Graph convolutional models are also well suited and have yielded high accuracies\cite{gcn, mage}.
Other approaches used 1D CNNs\cite{1dcnn}, or variants of recurrent neural networks\cite{cnng, ssrn}, and also probabilistic neural networks have been proposed\cite{pnn}.
\newline
However, ASD classification is not limited to fMRI data and there has been work using, for example, EEG\cite{eeg} or also more novel imaging approaches such as functional near-infrared spectroscopy\cite{fnirs}.

The current study aims to improve classification performance of ASD based on rs-fMRI data over the entire ABIDE~I dataset\cite{DiMartino2013} by leveraging the representational capabilities of modern transformer architectures.
We thus summarize our main contributions as follows:
\begin{enumerate}
    \item We propose a novel multi-atlas enhanced transformer framework for ASD classification using rs-fMRI data: METAFormer
    \item We demonstrate that self-supervised pretraining leads to significant improvements in performance without the requirement of additional data.
    \item We show that our model outperforms state of the art methods on the ABIDE I dataset.
\end{enumerate}

\section{Methods}
\subsection{Dataset}
Our experiments are conducted on the ABIDE I dataset \cite{DiMartino2013} which is a publicly available dataset containing structural MRI as well as rs-fMRI data obtained from individuals with Autism Spectrum Disorder (ASD) and typical controls (TC) from 17 different research sites.
The raw dataset encompasses a total of 1112 subjects, 539 of which are diagnosed with ASD and 573 are TC.
Subjects ages range from 7 to 64 years with a median age of 14.7 years across groups.
The ABIDE I dataset is regarded as one of the most comprehensive and widely used datasets in the field, offering a combination of MRI, rs-fMRI, and demographic data.\newline
The ABIDE I dataset exhibits significant heterogeneity and variations that should be taken into account.
It comprises data from diverse research sites worldwide, leading to variations in scanning protocols, age groups, and other relevant factors.
Consequently, the analysis and interpretation of the ABIDE I dataset pose challenges due this inherent heterogeneity.
\subsubsection{Preprocessing Pipeline.}
We utilize the ABIDE I dataset provided by the Preprocessed Connectomes Project (PCP) \cite{pcp} for our analysis.
The PCP provides data for ABIDE I using different preprocessing strategies.
In this work we use the preprocessed data from the DPARSF pipeline \cite{dparsf} comprising 406 ASD and 476 TC subjects.
The DPARSF pipeline is based on SPM8 and includes the following steps:
The first 4 volumes of each fMRI time series are discarded to allow for magnetization stabilization.
Slice timing correction is performed to correct for differences in acquisition time between slices.
The fMRI time series are then realigned to the first volume to correct for head motion.
Intensity normalization is not performed.
To clean confounding variation due to  physiological noise, 24-parameter head motion, mean white matter and CSF signals are regressed out.
Motion realignment parameters are also regressed out as well as linear and quadratic trends in low-frequency drifts.
Bandpass filtering was performed after regressing nuisance signals to remove high-frequency noise and low-frequency drifts.
Finally, functional to anatomical registration is performed using rigid body transformation and anatomical to standard space registration is performed using DARTEL\cite{dartel}.

\subsubsection{Functional Connectivity.}
As the dimensionality of the preprocessed data is very high, we perform dimensionality reduction by dividing the brain into a set of parcels or regions with similar properties according to a brain atlas.
In this work we process our data using three different atlases.
The first atlas is the Automated Anatomical Labeling (AAL) atlas \cite{aal}.
This atlas, which is widely used in the literature, divides the brain into 116 regions of interest (ROIs) based on anatomical landmarks and was fractionated to functional resolution of $3mm^3$ using nearest-neighbor interpolation.
The second atlas is the Craddock 200 (CC200) atlas \cite{craddock200}.
It divides the brain into 200 ROIs based on functional connectivity and was fractionated to functional resolution of $3mm^3$ using nearest-neighbor interpolation.
The third atlas we considered is the Dosenbach 160 (DOS160) atlas \cite{dosenbach160} which contains uniform spheres placed at coordinates obtained from meta-analyses of task-related fMRI studies.\newline
After obtaining the ROI time-series from the atlases, we compute the functional connectivity using the Pearson Correlation Coefficient between each pair of ROIs.
The upper triangular part of the correlation matrix as well as the diagonal are then dropped and the lower triangular part is vectorized to obtain a feature vector of length $k(k-1)/2$, where $k$ is the number of ROIs, which is then standardized and serves as input to our models.

\subsection{Model Architecture}
\subsubsection{METAFormer: Multi-Atlas Transformer.}
\begin{figure}[htbp]
\centerline{\includegraphics[width=\textwidth]{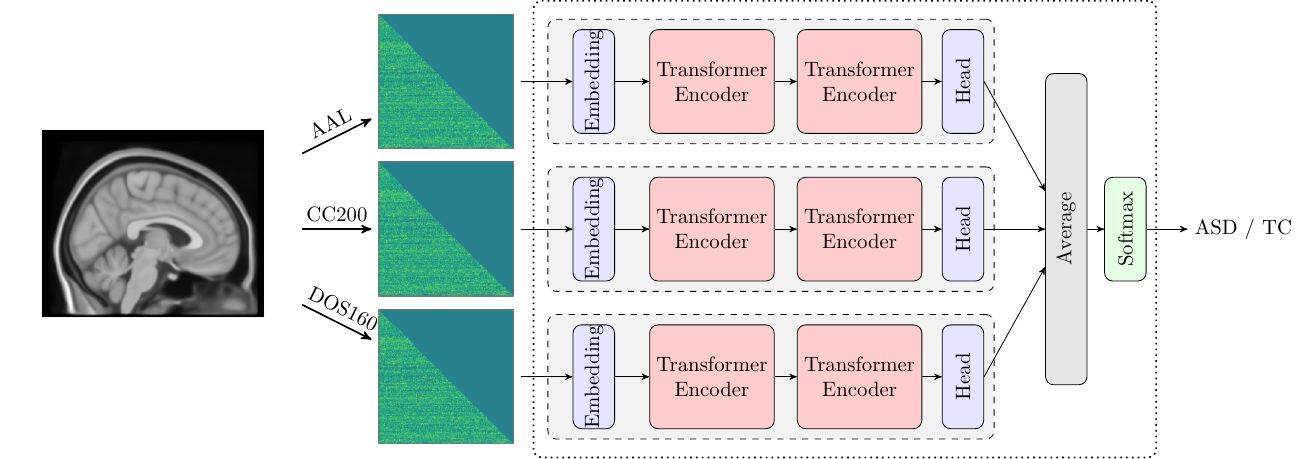}}
\caption{Architecture of METAFormer. Our model consists of three separate transformers, each corresponding to a specific atlas. The input to each transformer is a batch of flattened functional connectivity matrices with the diagonal and the upper triangular part of the matrix removed. The output of the transformers is averaged and passed through a softmax layer to derive the final class probabilities.}
\label{fig:transformer}
\end{figure}

Here, we propose METAFormer, at the core of which lies the transformer encoder architecture, originally proposed by Vaswani et al.\cite{transformer} for natural language processing tasks.
However, as our main goal is to perform classification and not generation we do not use the decoder part of the transformer architecture.
In order to accommodate input from multiple different atlases, we employ an ensemble of three separate transformers, with each transformer corresponding to a specific atlas.
As depicted in Figure \ref{fig:transformer}, the input to each transformer is a batch of flattened functional connectivity matrices.
First, the input to each transformer undergoes embedding into a latent space using a linear layer with a dimensionality of $d_{model}=256$.
The output of the embedding is then multiplied by $\sqrt{d_{model}}$ to scale the input features.
This scaling operation aids in balancing the impact of the input features with the attention mechanism.
Since we are not dealing with sequential data, positional encodings are not utilized.\newline
The embedded input is subsequently passed through a BERT-style encoder\cite{bert}, which consists of $N=2$ identical layers with $d_{ff}=128$ feed forward units, and $h=4$ attention heads.
To maintain stability during training, each encoder layer is normalized using layer normalization \cite{layernorm}, and GELU \cite{hendrycks2023gaussian} is used as the activation function.
Following the final encoder layer, the output passes through a dropout layer.
Then, a linear layer with $d_{model}$ hidden units and two output units corresponding to the two classes is applied to obtain the final output.
The outputs of the three separate transformers are averaged, and this averaged representation is passed through a softmax layer to derive the final class probabilities.\newline

To train our Multi-Atlas Transformer model, we follow a series of steps. Firstly, we initialize the model weights using the initialization strategy proposed by He \cite{he2015delving}, while setting the biases to 0. To optimize the model, we employ the AdamW optimizer \cite{adamW} and minimize the binary cross entropy between predictions and labels. Our training process consists of 750 epochs, utilizing a batch size of 256. To prevent overfitting, we implement early stopping with a patience of 40 epochs.
In order to ensure robustness of our model, we apply data augmentation. Specifically, we randomly introduce noise to each flattened connectome vector with an augmentation probability of 0.3. The noise standard deviation is set to 0.01.
We conduct hyperparameter tuning using grid search. We optimize hyperparameters related to the optimizer, such as learning rate and weight decay. We also consider the dropout rate during this process.

\subsection{Self-Supervised Pretraining}
\begin{figure}[htbp]
\centerline{\includegraphics[width=0.7\textwidth]{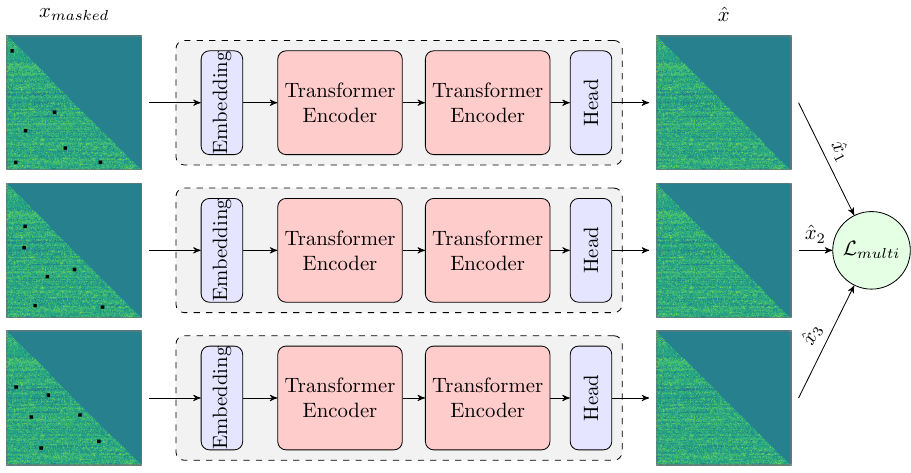}}
\caption{Self-supervised pretraining on the imputation task of the METAFormer architecture. The inputs to the model are masked connectomes, where 10\% of the features are randomly set to 0 (exemplified as black squares). The model is trained to predict the missing values implying that the output of the model has the same shape as the input.}
\label{fig:pretraining}
\end{figure}

As popularized by \cite{gpt}, the utilization of self-supervised generative pretraining followed by task-specific fine-tuning has demonstrated improved performance in transformer architectures.
Building upon this approach, we propose a self-supervised pretraining task for our model.
Our approach involves imputing missing elements in the functional connectivity matrices, drawing inspiration from the work introduced by \cite{mvts}.
To simulate missing data, we randomly set 10\% of the standardized features in each connectome to 0 and train the model to predict the missing values.
The corresponding configuration is illustrated in Figure \ref{fig:pretraining}.
To achieve this, we begin by randomly sampling a binary noise mask $M \in \{0,1\}^{n_i}$ for each training sample, where $n_i$ denotes the number of features in the $i$-th connectome.
Subsequently, the original input $X$ is masked using element-wise multiplication with the noise mask: $X_{masked} = X \odot M$.\newline
To estimate the corrupted input, we introduce a linear layer with $n_i$ output neurons on top of the encoder stack, which predicts $\hat{x}_i$.
We calculate a multi atlas masked mean squared error (MAMSE) loss $\mathcal{L}_{multi}$ between the predicted and the original input:
\begin{equation}
    \mathcal{L}_{multi} = \frac{1}{3} \sum_{i=1}^{3} \frac{1}{n_i} \sum_{j\in M}^{n_i} ||x_{i,j} - \hat{x}_{i,j}||^2
\end{equation}
where $x_{i,j}$ is the original value of the $j$-th masked input from the $i$-th atlas and $\hat{x}_{i,j}$ is the predicted value for the masked input at position $j$ in the $i$-th atlas.

\section{Experiments}
\subsection{Experimental Setup}
To evaluate the classification performance of our models in a robust manner, we employed 10-fold stratified cross-validation.
For each fold, the model is trained on the remaining nine training folds and evaluated on the held-out test fold.
Further, we set aside 30\% of each training fold as validation sets which are then used for hyperparameter tuning and early stopping.

In order to assess the impact of self-supervised pretraining, we compare the performance of our model with and without pretraining.
To achieve that, we first pretrain the model using the imputation task on the training folds and subsequently fine-tune the model on the training folds using the classification task after which we evaluate on the held-out test fold.

In order to verify the efficacy of using multiple atlases as input we compared the performance of our METAFormer model with the performance of single atlas transformer (SAT) models.
For that, we trained three separate transformer models using only one atlas as input.
The SAT models are trained using the same architecture as well as training procedure as the METAFormer model.
We also evaluated the performance of the SAT models with and without self-supervised pretraining in order to asses its impact on the performance of the model.
To make results comparable, we use the same training and validation folds for all model configurations under investigation.

\subsection{Evaluation Metrics}By using cross-validation, we obtained 10 different sets of performance scores per configuration.
These scores were then averaged and the standard deviation of each score was obtained, providing reliable estimates of the model's performance on unseen data.
The classification results were reported in terms of accuracy, precision, recall (sensitivity), F1-score and AUC-score, which are commonly used metrics for evaluating classification models.

\section{ASD Classification Results}
Table \ref{tab:comparison} shows the superior performance of our pretrained METAFormer model compared to previously published ASD classifiers trained on atlas-based connectomes.
Importantly, our model achieves higher accuracy even when compared to approaches with similar test set sizes that did not employ cross-validation.\newline
To further validate the effectiveness of our proposed Multi-Atlas Transformer model for Autism Spectrum Disorder classification, we compare METAFormer against single atlas transformers.
The results, as presented in Table \ref{tab:results}, demonstrate the superiority of METAFormer over all single atlas models in terms of accuracy, precision, recall, F1-score, and AUC-score.
Moreover, the multi-atlas model exhibits comparable or lower standard deviations in performance metrics compared to the single atlas models.
This indicates higher robustness and stability of our multi-atlas METAFormer architecture, attributed to the joint training of the three transformer encoders.
\begin{table}[h!]
    \centering
    \ra{1.3}
    \begin{tabular}{@{}lllllcr@{}}\toprule
        \textbf{Study} & \#\textbf{ASD} & \#\textbf{TC} & \textbf{Model} & \textbf{CV} & \textbf{Acc.} \\ \midrule
        MAGE\cite{mage} & 419 & 513 & Graph CNN & 10-fold& 75.9\% \\
        AIMAFE\cite{aimafe} & 419 & 513 & MLP & 10-fold& 74.5\% \\
        1DCNN-GRU\cite{1dcnn} & \ -- &\ -- & 1D CNN & \ --& 78.0\% \\
        MISODNN\cite{misodnn} & 506 & 532 & MLP & 10-fold & 78.1\% \\
        3D CNN\cite{ensemble} & 539 & 573 & 3D CNN & 5-fold& 74.53\% \\
        CNNGCN\cite{cnng} & 403 & 468 & CNN/GRU &\ --& 72.48\% \\
        SSRN\cite{ssrn} & 505 & 530 & LSTM & \ --& 81.1\% \\
        \textbf{Ours} & 408 & 476 & Transformer & 10-fold& \textbf{83.7\%} \\
        \bottomrule
    \end{tabular}
    \caption{Overview of state-of-the-art ASD classification methods that use large, heterogenous samples from ABIDE I. Note that our model achieves the highest accuracy while still using 10-fold cross-validation.}
    \label{tab:comparison}
\end{table}

\subsection{Impact of Pretraining}
We also evaluated the effect of self-supervised pretraining on the classification performance of our models.
As Table \ref{tab:results} shows pretraining significantly improves the performance of all models in terms of accuracy, precision, recall, F1-score and AUC-score.
Furthermore, for our proposed METAFormer architecture pretraining improves the performance by a large margin.
\begin{table*}[htbp]
\centering
\ra{1.2}
\begin{tabular}{@{}llllll@{}}\toprule
    \textbf{Variant} & \textbf{Acc.} & \textbf{Prec.} & \textbf{Rec.} & \textbf{F1} & \textbf{AUC}\\ \midrule
    METAFormer PT     & \textbf{0.837} \tiny{$\pm$0.030} & \textbf{0.819}\tiny{$\pm$0.045} & 0.901 \tiny{$\pm$0.044} & \textbf{0.856} \tiny{$\pm$0.023} & \textbf{0.832} \tiny{$\pm$0.032}\\ 
    METAFormer        & 0.628 \tiny{$\pm$0.041} & 0.648\tiny{$\pm$0.040} & 0.688 \tiny{$\pm$0.091} & 0.663 \tiny{$\pm$0.047} & 0.623 \tiny{$\pm$0.041}\\ \midrule
    SAT (AAL)       & 0.593 \tiny{$\pm$0.040} & 0.585\tiny{$\pm$0.042} & 0.888 \tiny{$\pm$0.091} & 0.701 \tiny{$\pm$0.024} & 0.568 \tiny{$\pm$0.047}\\
    SAT (CC200)     & 0.586 \tiny{$\pm$0.037} & 0.577\tiny{$\pm$0.027} & 0.888 \tiny{$\pm$0.057} & 0.698 \tiny{$\pm$0.019} & 0.560 \tiny{$\pm$0.044}\\
    SAT (DOS160)    & 0.570 \tiny{$\pm$0.055} & 0.571\tiny{$\pm$0.038} & 0.816 \tiny{$\pm$0.101} & 0.670 \tiny{$\pm$0.051} & 0.550 \tiny{$\pm$0.056} \\
    SAT (AAL) PT    & 0.601 \tiny{$\pm$0.069} & 0.587\tiny{$\pm$0.055} & 0.939 \tiny{$\pm$0.059} & 0.719 \tiny{$\pm$0.033} & 0.573 \tiny{$\pm$0.077}\\
    SAT (CC200) PT  & 0.632 \tiny{$\pm$0.071} & 0.622\tiny{$\pm$0.074} & 0.891 \tiny{$\pm$0.102} & 0.724 \tiny{$\pm$0.035} & 0.611 \tiny{$\pm$0.082}\\
    SAT (DOS160) PT & 0.683 \tiny{$\pm$0.094} & 0.652\tiny{$\pm$0.091} & \textbf{0.964} \tiny{$\pm$0.057} & 0.771 \tiny{$\pm$0.047} & 0.660 \tiny{$\pm$0.106} \\
    \bottomrule
\end{tabular}
\caption{Classification results for the different model configurations. Reported values are the mean $\pm$ standard deviation over 10 folds. The best results are in bold. SAT=Single Atlas Transformer, PT=Pretrained, atlases are in braces. Note that pretraining significantly improves performance across metrics and atlases. Using our multi-atlas METAFormer in combination with pretraining yields impressive performance increases.}
\label{tab:results}
\end{table*}
\section{Conclusion}
In this paper, we propose METAFormer, a novel multi-atlas enhanced pretrained transformer architecture for ASD classification.
We utilize self-supervised pretraining on the imputation task on the same dataset to prime the model for the downstream task.
We conducted extensive experiments to demonstrate the effectiveness of our approach by comparing it to several baselines that use single-atlas and multi-atlas approaches with and without pretraing.
Our results show that our model performs better than state-of-the-art methods and that pretraining is highly beneficial for the downstream task.

\section{Acknowledgements}
The authors thank the International Max Planck Research School for the Mechanisms of Mental Function and Dysfunction (IMPRS-MMFD) for supporting Samuel Heczko. Florian Birk is supported by the Deutsche Forschungsgesellschaft (DFG) Grant DFG HE 9297/1-1. Julius Steiglechner is funded by Alzheimer Forschung Initiative e.V.Grant \#18052.
\end{document}